\pdfoutput=1

\documentclass[11pt]{article}

\usepackage{acl}

\usepackage{times}
\usepackage{latexsym}

\usepackage[T1]{fontenc}

\usepackage[utf8]{inputenc}

\usepackage{microtype}

\usepackage{inconsolata}

\usepackage{graphicx}

\usepackage{coffeestains}

\usepackage{amsmath}
\usepackage{amssymb}
\usepackage{pifont}
\usepackage{booktabs}
\usepackage{algorithm}
\usepackage{algpseudocode}
\algrenewcommand\algorithmicrequire{\textbf{Input:}}
\algrenewcommand\algorithmicensure{\textbf{Output:}}
\usepackage{algpseudocode}
\usepackage{threeparttable}
\usepackage{multirow}
\usepackage{tabularx}
\usepackage{cleveref}
\usepackage{enumitem}
\usepackage[textsize=small,color=gray]{todonotes}
\usepackage{mathtools}
\usepackage{soul}
\usepackage{multirow}
\usepackage{colortbl}
\usepackage{tcolorbox}
\colorlet{lightgray}{White!30!lightgray}
\colorlet{lightblue}{White!70!MidnightBlue}

\crefname{figure}{Figure}{Figures}
\crefname{equation}{Equation}{Equations}
\crefname{appendix}{Appendix}{Appendices}
\crefname{table}{Table}{Tables}
\crefname{section}{\S}{\S\S}
\newcommand{\interalia}{\emph{inter alia}}

\title{Distillation versus Contrastive Learning: How to Train Your Rerankers}

\author{
Zhichao Xu $^{1 2}$
Zhiqi Huang $^{3}$
Shengyao Zhuang $^{4}$
Vivek Srikumar $^{1}$\\
$^{1}$ Kahlert School of Computing, University of Utah \quad \\
$^{2}$ Scientific Computing and Imaging Institute, University of Utah \quad \\
$^{3}$ Capital One, Inc \,\, $^{4}$ The University of Queensland \\
{\tt zhichao.xu@utah.edu \,\, svivek@cs.utah.edu}
}

\begin{document}
\maketitle

\begin{abstract}
Training effective text rerankers is crucial for information retrieval. Two strategies are widely used: contrastive learning (optimizing directly on ground-truth labels) and knowledge distillation (transferring knowledge from a larger reranker). 
While both have been studied extensively, a clear comparison of their effectiveness for training cross-encoder rerankers under practical conditions is needed.

This paper empirically compares these strategies by training rerankers of different sizes (0.5B, 1.5B, 3B, 7B) and architectures (Transformer, Recurrent) using both methods on the same data, with a strong contrastive learning model acting as the distillation teacher. Our results show that knowledge distillation generally yields better in-domain and out-of-domain ranking performance than contrastive learning when distilling from a more performant teacher model. This finding is consistent across student model sizes and architectures. However, distilling from a teacher of the same capacity does not provide the same advantage, particularly for out-of-domain tasks. These findings offer practical guidance for choosing a training strategy based on available teacher models. We recommend using knowledge distillation to train smaller rerankers if a larger, more performant teacher is accessible; in its absence, contrastive learning remains a robust baseline. Our code implementation is made available to facilitate reproducbility.\footnote{\url{https://github.com/zhichaoxu-shufe/distillation-versus-contrastive-learning}}
\end{abstract}

\section{Introduction}
\label{sec:intro}
Modern information retrieval (IR) systems often rely on a two-stage process: an initial retriever quickly finds candidate texts, and a more powerful reranker re-orders them by relevance to improve the final ranking quality~\cite[\interalia]{schutze2008introduction,ma2023fine,asai2024openscholar,singh2025ai2scholarqa}. This reranking stage, which involves scoring and sorting texts by their relevance to a given query, commonly employs powerful cross-encoders~\cite{yates2021pretrained,zhuang2024setwiseapproachforeffectiveandhighlyefficient}.

Effective training is key to building a high-performing cross-encoder pointwise reranker. Two main strategies have emerged for this purpose. The first, contrastive learning (CL), trains the model directly using ground-truth relevance labels, learning to distinguish positive (relevant) examples from negative (irrelevant) ones~\cite{oord2018representation,gao2021rethink,ma2023fine}. The second, knowledge distillation (KD), instead involves training a smaller ``student'' model to replicate the outputs of a larger, more capable ``teacher'' model~\cite{buciluǎ2006modelcompression,hinton2015distillingknowledge,hofstatter2020marginmse,schlatt2025rankdistillm}. This method is often employed to create efficient models that can approximate the performance of larger ones.

Both strategies are widely used to train rerankers. But for practical deployments, is one preferable to the other? Prior work has explored related, but distinct, questions. For instance, \citet{hofstatter2020marginmse} explored distilling from rerankers to retrievers; while \citet{baldelli2024twolaratwostepllmaugmenteddistillation} and~\citet{schlatt2025rankdistillm} focused on training rerankers via distilling from large language models (LLMs) like GPT-4~\cite{achiam2023gpt4techinicalreport}. A line of works empirically compared distillation versus contrastive learning in training learning-to-rank models and BERT-sized rankers~\cite{qin2021improvingneuralrankingvialosslessknowledgedistillation,qin2023rdsuite,qin2021neuralrankers,zhuang2021ensembledistillation}. However, there lacks a direct comparison of these two strategies in the context of modern rerankers based on casual language models~\cite{ma2023fine,xu-etal-2025-statespacemodelsarestrongtextrerankers}.

In this paper, we directly bridge this gap by presenting an empirical comparison of contrastive learning and knowledge distillation as training strategies for cross-encoder rerankers.
To this end, we conduct a suite of controlled experiments. We train cross-encoder models of various sizes (0.5B, 1.5B, 3B, 7B) and different architectures (Transformer and Recurrent) using both strategies. For knowledge distillation, we employ a performant 7B Qwen2.5 model~\cite{yang2024qwen2.5technicalreport} trained with contrastive learning as the teacher. We focus on modern decoder-only models, as recent works have reported their effectiveness over smaller encoder-based models like BERT~\cite[\interalia]{ma2023fine,muennighoff2024gritlm,zhang2025qwen3embedding}. We train all models on the same dataset and evaluate their ranking performance on standard in-domain and out-of-domain benchmarks.

Our findings consistently show that training with knowledge distillation yields better ranking performance than direct contrastive learning when the student model is smaller than the teacher. For instance, distilling from our 7B teacher significantly improved the performance of 0.5B, 1.5B, and 3B student models on both in-domain and out-of-domain benchmarks compared to training them with contrastive learning. This advantage holds across both Transformer and Recurrent model architectures. However, this benefit diminishes when the student and teacher have the same capacity; distilling from a 7B teacher to a 7B student yielded no significant improvements and, in some cases, harmed out-of-domain generalization~\cite{gou2021knowledgedistillationsurvey,Gholami2023canastudentllmperformaswellasteacher}.
To further test the robustness of our conclusions, we conduct an additional experiment where models are trained on a diverse, multi-domain dataset RLHN~\citep{thakur2025fixingdatathathurtsperformance} instead of only on hard negatives mined from a single source. In this setting, knowledge distillation again proves to be the more effective strategy, confirming its advantage even when the reranker's training data is decoupled from the first-stage retriever.
Together, these findings position knowledge distillation as a powerful and robust strategy for training smaller rerankers, provided a performant teacher is available. In its absence, we find contrastive learning remains a robust baseline.

\section{Related Works}
\label{sec:related}
In this paper, we focus exclusively on the architecture of cross-encoder pointwise reranker~\cite{nogueira2019passage,nogueira2020document,zhuang2024setwiseapproachforeffectiveandhighlyefficient}. The reranker model processes a query and a document simultaneously, allowing the self-attention mechanism to explicitly model the interactions between their tokens throughout the layers.
\citet{nogueira2019passage} first demonstrated the efficacy of this approach by fine-tuning BERT as a text pair classifier. 
The fine-tuning of the BERT-type model as cross-encoder reranker has later been extended to encoder-decoder architectures, such as T5~\cite{raffel2020exploring,nogueira2020document}, and decoder-only models like Llama~\cite{touvron2023llama,ma2023fine}.

The training objective for a cross-encoder is to accurately discriminate between relevant and non-relevant documents for a given query. In practice, the non-relevant documents are usually hard negatives mined from first-stage retrievers~\cite[\interalia]{nogueira2019document,gao2021rethink,boytsov2022understanding}, and optionally combined with synthetic data curated from a complex and scalable data pipeline~\cite{zhang2025qwen3embedding,wang2025jinarerankerv3}. In literature, two primary training strategies are widely used to train cross-encoder pointwise reranker. The first is contrastive learning, which trains the model directly on ground-truth labels to maximize the distinction between positive and negative examples~\cite{gao2021rethink,zhuang2023rankt5}. The second strategy is knowledge distillation, which involves training a student reranker to mimic the behavior and performance of a larger, more capable teacher model~\cite{hofstatter2020marginmse,baldelli2024twolaratwostepllmaugmenteddistillation,schlatt2025rankdistillm}. 
Knowledge distillation~\cite[KD,][]{buciluǎ2006modelcompression,hinton2015distillingknowledge} facilitates the transfer of knowledge from a large, complex ``teacher'' model to a smaller, more efficient ``student'' model. The goal is to enable the student to mimic the teacher's output at the logit level, thereby inheriting its predictive capabilities at a reduced computational cost. Within the context of training neural IR models, knowledge distillation has been effectively applied to enhance both bi-encoder retriever~\cite[\interalia]{hofstatter2020marginmse,formal2021spladev2sparselexicalandexpansionmodel} and cross-encoder reranker~\cite{baldelli2024twolaratwostepllmaugmenteddistillation,schlatt2025rankdistillm}.\footnote{Note that, we use a narrower definition of knowledge distillation that focuses on aligning the student's output with the teacher's at the logit level~\cite{hinton2015distillingknowledge}. Sequence-level distillation~\cite{kim2016sequencelevel}, and training with synthetic data annotated by LLMs~\cite{zhang2025qwen3embedding,schlatt2025rankdistillm} fall into the broader scope of knowledge distillation.}
In this work, we aim to compare these two training strategies in a controlled experimental setting to evaluate their strengths and weaknesses.

\section{Methodology}
\label{sec:methodology}
Training strategy is critical for cross-encoder reranker's ranking performance. Two primary strategies have been extensively studied in the literature: (1) direct optimization on ground-truth labels via contrastive learning~\cite{gao2021rethink,yates2021pretrained,ma2023fine}, and (2) knowledge transfer from a larger model via knowledge distillation~\cite[\interalia]{hofstatter2020marginmse,schlatt2025rankdistillm}. 
\begin{tcolorbox}[colback=gray!5!white, colframe=black!75!black, boxrule=0.1mm, width=0.48\textwidth, arc=1mm, auto outer arc=true]
\textbf{Objective:} This paper aims to empirically compare contrastive learning and knowledge distillation for training cross-encoder rerankers. We aim to elucidate their respective strengths and provide clear guidance on which strategy is preferable under different practical constraints.
\end{tcolorbox}
\noindent
In the rest of this section, we first formally define the text reranking problem and our notation (\cref{subsec:textreranking}). We then provide detailed technical descriptions of the training process using contrastive learning (\cref{subsec:method_cl}) and knowledge distillation (\cref{subsec:method_kd}).
\subsection{The Text Reranking Problem}
\label{subsec:textreranking}
Modern IR systems often employ a two-stage retrieval-and-rerank pipeline~\cite[\interalia]{schutze2008introduction,zhang-etal-2021-learning-rank,asai2024openscholar}. 
An efficient first-stage retriever initially fetches a broad set of candidate texts. Subsequently, a more powerful reranker refines this initial list to optimize ranking metrics. Reranking is the task of ordering texts (e.g., passages or documents) by their relevance to a given query.

Let $q$ be an input query, and $d$ be a text from a corpus $\mathcal{D}$.
We define a reranking model $f_{\theta}(q,d)$, parameterized by $\theta$, which computes a scalar relevance score.
This model is typically a cross-encoder: the query $q$ and text $d$ are concatenated and fed into a transformer-based language model, whose output is passed through a linear layer to produce the final score~\cite[\interalia]{yates2021pretrained,nogueira2019document,boytsov2022understanding,ma2023fine,xu2024rankmamba,xu2025surveyofmodelarchitectures}.

\subsection{Training with Contrastive Learning}
\label{subsec:method_cl}
Contrastive learning is one strategy for learning representations~\cite{oord2018representation,weng2021contrastive}. 
Its application to reranking builds on the principle of the InfoNCE loss~\cite{oord2018representation}, which is derived from Noise-Contrastive Estimation~\cite{gutmann10noisecontrastiveestimation}.

The general goal is to learn a model that distinguishes a ``positive'' data sample from a set of ``negative'' (or noise) samples, given a certain context. 
Denote a context vector $\mathbf{c}$, and consider a set of $N$ samples $X=\{\mathbf{x}_i\}_{i=1}^N$, where one sample $\mathbf{x}_{\text{pos}}$ is a positive sample drawn from the conditional distribution $p(\mathbf{x}|\mathbf{c})$, and the $N-1$ negative samples are drawn from a proposal distribution $p(\mathbf{x})$. Using Bayes' rule, the probability that a sample $\mathbf{x}_i$ is the positive one is:
\begin{align*}
p(C = \text{pos}|X, \mathbf{c}) =
\frac{p(\mathbf{x}_{\text{pos}}|\mathbf{c}) \prod_{i \neq \text{pos}} p(\mathbf{x}_i)}
{\sum_{j} \left[ p(\mathbf{x}_j|\mathbf{c}) \prod_{i \neq j} p(\mathbf{x}_i) \right]} \\
= \frac{p(\mathbf{x}_{\text{pos}}|\mathbf{c}) / p(\mathbf{x}_{\text{pos}})}
{\sum_{j} p(\mathbf{x}_j|\mathbf{c}) / p(\mathbf{x}_j)}
= \frac{f(\mathbf{x}_{\text{pos}}, \mathbf{c})}
{\sum_{j} f(\mathbf{x}_j, \mathbf{c})}
\end{align*}
We can define the scoring function that is proportional to the density ratio $f(\mathbf{x},\mathbf{c})\propto\frac{p(\mathbf{x}, \mathbf{c})}{p(\mathbf{c})}$. The InfoNCE loss optimizes the negative log probability of classifying the positive sample correctly:
\begin{equation}
    \mathcal{L}_{\text{InfoNCE}}=-\mathbb{E}\big[ \log \frac{f(\mathbf{x}, \mathbf{c})}{\sum_{\mathbf{x'}\in \mathbf{X}}f(\mathbf{x'}, c)}\big]
\end{equation}
We map the abstract concepts to our reranking task:
\begin{itemize}
    \item The context $\mathbf{c}$ is the query $q_i$.
    \item The positive sample $\mathbf{x}_{\text{pos}}$ is the relevant document $d_i^+$.
    \item The negative samples $\{\mathbf{x}_k\}$ are a set of irrelevant documents $D_i^-$.
    \item The scoring function $f$ is the parameterized reranker model $f_{\theta}$.
\end{itemize}
For a training instance consisting of a query $q_i$, a positive document $d_i^+$, and a set of negative documents $D_i^-$, the loss is:
$$
    \resizebox{0.48\textwidth}{!}{$-\frac{1}{|\mathcal{S}|} \sum\limits_{(q_i, d_i^{+})\in \mathcal{S}} \log  \frac{\exp f_{\theta}(q_i, d_i^{+})}{\exp f_{\theta}(q_i, d_i^{+}) + \sum\limits_{j\in \mathcal{D}_i^{-}} \exp f_{\theta}(q_i, d_i^{-})}$}
$$
The total loss is averaged over all training instances. Following common practice~\cite{gao2021rethink,xu2024rankmamba}, the negative documents are often "hard negatives"\,---\,documents that the first-stage retriever ranked highly but are not labeled as relevant. In practice, training instances are grouped into minibatches, and the parameters $\theta$ are optimized jointly.

\subsection{Training with Knowledge Distillation}
\label{subsec:method_kd}
Knowledge distillation (KD) is a technique for training a smaller, efficient "student" model by transferring knowledge from a larger, more capable "teacher" model~\cite{hinton2015distillingknowledge,gou2021knowledgedistillationsurvey}. In IR, KD is used to create fast rerankers that approximate the performance of slower, larger models, which is critical for production systems~\citep[\interalia]{hofstatter2020marginmse,santhanam-etal-2022-colbertv2}.

Let $f_t$ denote the teacher reranker and $f_s$ denote the student reranker. For a given query $q$ and a list of candidate texts $\mathcal{D}_q=\{d_1, d_2, \ldots, d_k\}$, we first compute relevance scores from both models. This yields two score vectors (logits):
\begin{align*}
\mathbf{z}_t &= [f_t(q, d_1), f_t(q, d_2), \dots, f_t(q, d_k)] \\
\mathbf{z}_s &= [f_s(q, d_1), f_s(q, d_2), \dots, f_s(q, d_k)]
\end{align*}
The student model $f_s$ is trained to mimic the teacher's output distribution over the candidate texts. This is achieved by minimizing the Kullback-Leibler (KL) divergence between the two softened probability distributions:
$$
\mathcal{L}_{\text{KD}} = D_{\mathrm{KL}}\left( \mathrm{softmax}\left(\frac{\mathbf{z}_t}{T}\right) \bigg| \mathrm{softmax}\left(\frac{\mathbf{z}_s}{T}\right) \right)
$$
where $T$ is the temperature hyperparameter. A higher temperature creates a softer probability distribution, which can help in transferring more nuanced information from the teacher. In practice, $T$ is often set to 1~\cite{hinton2015distillingknowledge}.

Whereas contrastive learning optimizes a model on ground-truth labels, knowledge distillation mimics the outputs of a more capable teacher. The central goal of this paper is to empirically compare these distinct paradigms for training cross-encoder rerankers under controlled settings, which we detail in the following section.

\section{Experimental Setup}
\label{sec:experiments}
\paragraph{Training setup.} 
We construct our training set based on the well established MS MARCO passage retrieval dataset~\citep{bajaj2016ms}. The official training set contains 532k training pairs, and the corpus contains 8.8M passages. 
For first stage retriever, we reproduce the RepLlama experiment~\cite{ma2023fine}, but replace the Llama-2-7B backbone model~\cite{touvron2023llama} with stronger Qwen2.5-7B model~\cite{yang2024qwen2.5technicalreport}. The retriever\,---\,named RepQwen\,---\,is trained on \texttt{Tevatron/MSMARCO-passage-aug}\footnote{\url{https://huggingface.co/datasets/Tevatron/msmarco-passage-aug}} trainset, which consists of 486k $(q_i, d_i^+)$ pairs with hard negatives mined from BM25 and CoCondenser~\cite{gao2022unsupervised}.

To construct the training set for the rerankers, we follow the same strategy as~\citet{ma2023fine} to mine hard negatives from the first stage retriever. Specifically, for each $(q_i, d_i^+)$ pair, we randomly sample $k$ negatives from top-200 passages retrieved by the retriever, excluding $d_i^+$. As shown in the experimental results (\cref{sec:results}), the models trained using this constructed dataset achieves on par or improved performance over RankLlama2~\cite{ma2023fine}, suggesting the correctness of the pipeline. We did not further investigate hard negative mining strategies like prior works~\cite{yu2024arcticembed2.0,thakur2025fixingdatathathurtsperformance,lee2025nvembed} as they are orthogonal to the goal of this paper.

The $(q_i, d_i^+)$ pairs, together with the mined hard negatives are then used to train rerankers with contrastive learning. We first train the most capable reranker\,---\,named RankQwen-7B\,---\,with Qwen2.5-7B backbone using contrastive learning. This RankQwen-7B is then used as the teacher model to label $\mathcal{D}_i=\{d_i^+, D_i^-\}$ to be subsequently used to train student rerankers. This way, we create a controlled experiment as the rerankers trained with contrastive learning and knowledge distillation are trained on the same $(q, d, \mathcal{D}^-)$ triples, and the only difference is the training strategy. 

\paragraph{Evaluation setup.}
We evaluate the reranking performance following prior works' practices~\cite[\interalia]{ma2023fine,xu-etal-2025-statespacemodelsarestrongtextrerankers}. For in-domain evaluation, we use MS MARCO passage dev set, which includes 6,980 queries. We also include TREC DL19 and DL20~\cite{craswell2020overview,craswell2021overview} consisting of 43 and 54 queries respectively. 
For out-of-domain evaluation, we adopt BEIR benchmark~\cite{thakurbeir}. We evaluate performances on 13 subsets that are publicly available. Refer to~\cref{asec:licenses} for details of datasets.

We report the official evaluation metrics for all benchmarks, i.e., MRR@10 for MS MARCO passage dev, NDCG@10 for DL19 and DL20, NDCG@10 for BEIR benchmarks. 

\paragraph{Models used.}
Our main experiments are based on RepQwen, the retriever with a Qwen2.5-7B backbone following the RepLlama~\cite{ma2023fine} training strategy. We also report the performance of a retriever with a Llama-3.1-8B backbone. 

The objects of our study are reranker models. We evaluate 
Qwen2.5 models of different sizes, including 0.5B, 1.3B, 3B and 7B, to allow us to observe the scaling trend. For each model size, we train two reranker models: one via direct contrastive learning and one via distillation; we denote with CL/KD suffix, e.g., RankQwen-0.5B-CL means 0.5B model trained with contrastive learning. For knowledge distillation training, we distil from the RankQwen-7B-CL teacher.

We also experiment with RecurrentGemma~\cite{de2024griffin}, a recurrent language model based on the Griffin architecture instead of quadratic complexity Transformers~\cite{vaswani2017attention}. We use the 2B variant and refer to the trained model as RankRGemma-2B. Recent works have explored recurrent language models' efficacy for IR tasks, such as state space models like Mamba~\cite{gu2023mamba,dao2024transformers} for retrieval~\cite{zhang2024mamba} and reranking~\cite{xu2024rankmamba,xu-etal-2025-statespacemodelsarestrongtextrerankers}. We follow this direction to examine recurrent language models' efficacy under different training strategies. 

For all the models used in our experiments, we use the pretrained base models.

\paragraph{Baselines.}
Our baselines are prior results under the same training setting. RepLlama and RankLlama~\cite{ma2023fine} are the closest baselines which use the same trainset for contrastive learning, using Llama-2 backbone~\cite{touvron2023llama}. CSPLADE~\cite{xu2025csplade} is a learned sparse retrieval model with Llama-3-8B backbone~\cite{dubey2024llama}, and achieves competitive performance compared to dense retrieval models.
RankMamba-2~\cite{xu-etal-2025-statespacemodelsarestrongtextrerankers} trains Mamba-2-based rerankers for passage reranking, though their results are based on BGE retriever~\cite{bge_embedding}. We compare against the original numbers reported by the authors.

\paragraph{Implementation details.}
Our implementation is based on packages including PyTorch, Huggingface Transformers and Tevatron-v2~\cite{ma2025tevatron}. For all our models, we train with LoRA~\cite{hu2021lora} to balance in-domain and out-of-domain performance and reduce overfitting. For scalable training, we use DeepSpeed stage 2~\cite{aminabadi2022deepspeed}, activation checkpointing, FlashAttention-2~\cite{dao2024flashattention}, mixed precision and gradient accumulation. \Cref{asec:hyperparameters} gives details about hyperparameters.

\section{Results and Analysis}
\label{sec:results}
\subsection{Main Results}
\label{subsec:main_results}
\begin{table*}[h!]
\vspace{0pt}
\centering
\caption{Results for passage reranking in-domain evaluation. We mark best results in each section bold; $\dag$ indicates the overall best result and $\ddag$ indicates the best result among our trained models; for models in \emph{Retrieval} and \emph{Reranking} baseline sections, $^\clubsuit$ denotes the model we trained. Note RankQwen-7B-CL$^\clubsuit$ is used as the teacher for distillation.
}
\label{tab:results_passage}
\resizebox{0.95\textwidth}{!}{
\begin{tabular}{
lrrrrrr
}
\toprule
Model & Size & \multicolumn{2}{c}{Source} & DEV & DL19 & DL20 \\
\, & \, & prev. & top-$k$ & MRR@10 & \multicolumn{2}{c}{NDCG@10} \\
\midrule
\multicolumn{7}{c}{\emph{Retrieval}} \\
BM25~\cite{lin2021pyserini} & - & - & $|\mathcal{D}|$ & 18.4 & 50.6 & 48.0 \\
CoCondenser~\cite{gao2022unsupervised} & 110M & - & $|\mathcal{D}|$ & 38.2 & 71.7 & 68.4\\
RepLlama2~\cite{ma2023fine} & 7B & - & $|\mathcal{D}|$ & 41.2 & 74.3 & 72.1 \\
CSPLADE~\cite{xu2025csplade} & 8B & - & $|\mathcal{D}|$ & 41.3 & 74.1 & \textbf{72.8} \\
RepQwen$^\clubsuit$ & 7B & - & $|\mathcal{D}|$ & 42.2 & 73.2 & 72.5 \\
RepLlama3$^\clubsuit$ & 8B & - & $|\mathcal{D}|$ & \textbf{42.6} & \textbf{74.4} & \textbf{72.8}\\
\midrule
\multicolumn{7}{c}{\emph{Reranking}} \\
cross-SimLM~\cite{wang2022simlm} & 110M & bi-SimLM & 200 & 43.7 & 74.6 & 72.7 \\
RankT5~\cite{zhuang2023rankt5} & 220M & GTR & 1000 & 43.4 & - & -\\
RankMamba~\cite{xu-etal-2025-statespacemodelsarestrongtextrerankers} & 1.3B & BGE & 100 & 38.6 & 75.8 & 74.0 \\
RankLlama-7B~\cite{ma2023fine} & 7B & RepLlama2 & 200 & \textbf{44.9}$\dag$ & 75.6 & \textbf{77.4}$\dag$ \\
RankQwen-7B-CL$^\clubsuit$ & 7B & RepQwen & 200 & 44.8$\ddag$ & \textbf{77.4} & 77.1\\
\midrule
\multicolumn{7}{c}{\emph{Contrastive Learning}} \\
RankQwen-0.5B-CL & 0.5B & RepQwen & 200 & 42.1 & 75.7 & 72.9 \\
RankQwen-1.5B-CL & 1.5B & RepQwen & 200 & 43.5 & 75.8 & \textbf{75.4} \\
RankQwen-3B-CL & 3B & RepQwen & 200 & \textbf{43.9} & \textbf{76.8} & \textbf{75.4} \\
RankRGemma-2B-CL & 2B & RepQwen & 200 & 43.0 & 76.0 & 74.7 \\
\midrule
\multicolumn{7}{c}{\emph{Knowledge Distillation}} \\
RankQwen-0.5B-KD & 0.5B & RepQwen & 200 & 43.5 & 76.1 & 75.5 \\
RankQwen-1.5B-KD & 1.5B & RepQwen & 200 & 43.9 & 76.1 & 76.8 \\
RankQwen-3B-KD & 3B & RepQwen & 200 & \textbf{44.7} & 77.4 & \textbf{77.1}$\ddag$ \\
RankQwen-7B-KD & 7B & RepQwen & 200 & \textbf{44.7} & \textbf{77.5}$\dag$$\ddag$ & 77.0 \\
RankRGemma-2B-KD & 2B & RepQwen & 200 & 43.6 & 76.1 & 75.0 \\
\bottomrule
\end{tabular}
}
\vspace{0pt}
\end{table*}
\paragraph{In-domain results.}
\Cref{tab:results_passage} reports the reranking performance on MS MARCO Dev and DL19+DL20. 
We observe our trained retrieval models, i.e., RepLlama3 and RepQwen performs on par or better than RepLlama2 and CSPLADE, the two models trained with same trainset. Similarly, RankQwen-7B-CL is comparable to RankLlama2-7B, suggesting the correctness of our training pipeline. 

We now compare performances between rerankers trained with contrastive learning, and rerankers trained with knowledge distillation, with RankQwen-7B-CL as teacher. 
We notice that with the same model sizes, knowledge distillation achieves better ranking performance compared to contrastive learning. For example, RankQwen-0.5B-KD achieves 43.5 MRR@10 on Dev, and average 75.8 NDCG@10 on DL19+20's 97 queries, improving over RankQwen-0.5B-CL's 42.1 and 74.1. The similar observation applies for 1.5B and 3B scale Qwen models as well as RankRGemma\,---\,the recurrent model with Griffin architecture. These observations suggest the efficacy of knowledge distillation: the performance improvement is consistent across different model sizes (0.5B, 1.5B, 3B) and model architectures (Transformer, Griffin). 

We also note RankQwen-7B-KD achieves similar in-domain performance as RankQwen-3B-KD and the RankQwen-7B-CL teacher, which suggests that the student model is not benefiting from knowledge distillation training when the teacher model is of the same capacity. We will revisit this problem in our discussion of out-of-domain results.

\begin{table*}[ht]
\vspace{0pt}
\centering
\caption{Results for passage reranking out-of-domain evaluation. 0.5B, 1.5B, 3B, 7B 2B correspond to RankQwen-0.5B, RankQwen-1.5B, RankQwen-3B, RankQwen-7B and RankRGemma-2B, respectively. We also report results without Quora dataset as Quora duplicate question detection is a symmetric retrieval task, not aligning with the asymmetric web search task of the training set. We mark best results in each section bold; $\dag$ indicates the overall best result including the teacher model, $\ddag$ indicates the best result excluding teacher. }
\label{tab:results_beir}
\resizebox{0.95\linewidth}{!}{
\begin{tabular}{@{}l|r|rrrr|rrrrr@{}}
\toprule
\multicolumn{1}{l|}{\,}
& \multicolumn{1}{l|}{Teacher} 
& \multicolumn{4}{c|}{Contrastive Learning} 
& \multicolumn{5}{c}{Knowledge Distillation} \\
\multicolumn{1}{l|}{Dataset} 
& \multicolumn{1}{r|}{7B} & \multicolumn{1}{r}{0.5B} & \multicolumn{1}{r}{1.5B} & \multicolumn{1}{r}{3B} & \multicolumn{1}{r|}{2B} & \multicolumn{1}{r}{0.5B} & \multicolumn{1}{r}{1.5B} & \multicolumn{1}{r}{3B} & \multicolumn{1}{r}{7B} & \multicolumn{1}{r}{2B} \\ \midrule
\multicolumn{1}{l|}{Arguana} & 55.9 & 50.6 & 55.8 & \textbf{56.2} & 52.3 & 51.3 & \textbf{57.6}$\dag$$\ddag$ & 56.5 & 55.2 & 54.7 \\
\multicolumn{1}{l|}{Climate-FEVER} & 27.1 & 23.8 & 27.6 & \textbf{30.5}$\dag$$\ddag$ & 24.3 & 23.8 & \textbf{28.5} & 31.6 & 28.7 & 26.6 \\
\multicolumn{1}{l|}{DBPedia} & 48.7$\dag$ & 44.5 & 46.3 & \textbf{48.5}$\ddag$ & 47.0 & 46.4 & 47.7 & \textbf{48.5}$\ddag$ & 48.1 & 47.6 \\
\multicolumn{1}{l|}{FEVER} & 88.2 & 86.0 & 86.2 & \textbf{89.1}$\dag$ & 85.1 & 86.7 & 87.3 & \textbf{88.1} & 87.3 & 86.6 \\
\multicolumn{1}{l|}{FiQA} & 45.9 & 37.9 & 42.2 & 43.2 & \textbf{44.1} & 34.7 & 45.4 & \textbf{46.6}$\dag$$\ddag$ & 44.2 & 44.7 \\
\multicolumn{1}{l|}{HotpotQA} & 76.1 & 72.5 & 74.8 & \textbf{76.2} & 74.4 & 74.4 & 76.2 & \textbf{76.7}$\dag$$\ddag$ & 72.5 & 76.6 \\
\multicolumn{1}{l|}{NFCorpus} & 31.5 & 33.0 & 29.3 & 25.9 & \textbf{34.4}$\dag$$\ddag$ & 32.2 & 32.3 & 33.3 & 32.3 & \textbf{34.2} \\
\multicolumn{1}{l|}{NQ} & 65.9$\dag$ & 59.2 & 63.2 & \textbf{64.9} & 62.8 & 60.9 & 64.4 & \textbf{65.5}$\ddag$ & 64.2 & 64.2 \\
\multicolumn{1}{l|}{Quora} & 79.3 & 82.3 & \textbf{84.9}$\dag$$\ddag$ & 84.6 & 79.6 & 80.2 & 81.0 & \textbf{81.9} & 78.5 & 80.3 \\
\multicolumn{1}{l|}{SCIDOCS} & 19.1 & 16.9 & 16.2 & 17.3 & \textbf{18.4} & 17.8 & \textbf{19.2}$\dag$ & \textbf{19.2}$\dag$$\ddag$ & 19.1 & 18.6 \\
\multicolumn{1}{l|}{SciFact} & 75.0 & 72.2 & 72.1 & 71.1 & \textbf{73.7} & 75.6 & \textbf{75.8}$\dag$$\ddag$ & 75.5 & 74.6 & 75.2 \\
\multicolumn{1}{l|}{TREC-COVID} & 85.0 & 81.9 & 86.3 & \textbf{86.8} & 85.1 & 80.7 & 85.6 & 86.3 & 84.7 & \textbf{87.4}$\dag$$\ddag$ \\
\multicolumn{1}{l|}{Touche-2020} & 36.7 & 31.9 & 34.6 & \textbf{36.5} & 36.4 & 25.9 & 35.9 & \textbf{38.6}$\dag$$\ddag$ & 35.5 & 37.8 \\ \midrule
Average & 56.5 & 53.3 & 55.3 & \textbf{56.2} & 55.2 & 53.1 & 56.7 & \textbf{57.6}$\dag$$\ddag$ & 55.8 & 56.5 \\
Average w/o Quora & 56.2$\dag$ & 50.9 & 52.9 & \textbf{53.8} & 53.2 & 50.8 & 54.6 & \textbf{55.5}$\ddag$ & 53.9 & 54.5 \\ \bottomrule
\end{tabular}
}

\vspace{0pt}
\end{table*}

\paragraph{Out-of-domain results.}
We report the out-of-domain evaluation results in~\cref{tab:results_beir}. The baseline methods' results are deferred to~\cref{asec:baseline_results}.

We note similar observations as in the in-domain setting. 
Directly comparing contrastive learning to knowledge distillation, RankQwen-0.5B-KD achieves similar performance as RankQwen-0.5B-CL (53.1 average NDCG@10 v.s. 53.3); while RankQwen-1.5B-KD, RankQwen-3B-KD, RankRGemma-2B have better performance over their contrastive learning counterparts. 

Among all models with KD training strategy, RankQwen-3B-KD achieves strong out-of-domain performance, averaging 57.5 NDCG@10 over 13 BEIR datasets, improving over RankQwen-7B-CL teacher. However, we notice that RankQwen-7B-CL's underperformance is mainly due to Quora duplicate question retrieval\,---\,a symmetric retrieval task different from the asymmetric MS MARCO trainset. The performance is 55.5 versus 56.2 excluding Quora, suggesting that the student model still cannot improve over the teacher~\cite{gou2021knowledgedistillationsurvey,Gholami2023canastudentllmperformaswellasteacher}.

An important observation is that RankQwen-7B-KD leads to performance degradation in out-of-domain evaluation (55.8 vs RankQwen-3B-KD's 57.6), while their performances on in-domain MS MARCO datasets are comparable.
Prior studies have also noted that knowledge distillation suffers from overfitting and poor out-of-domain generalization~\cite{gou2021knowledgedistillationsurvey,Yuan2020Revisitingknowledgedistillation,Yun2020Regularizingclasswisepredictionsviaselfknowledgedistillation}.
Our experimental results suggest that when distilling from a larger, more capable teacher models (RankQwen-7B-CL in our case), the student still achieve performance improvement compared to contrastive training from scratch. 

\begin{figure*}[htp]
    \centering
    \includegraphics[width=\textwidth]{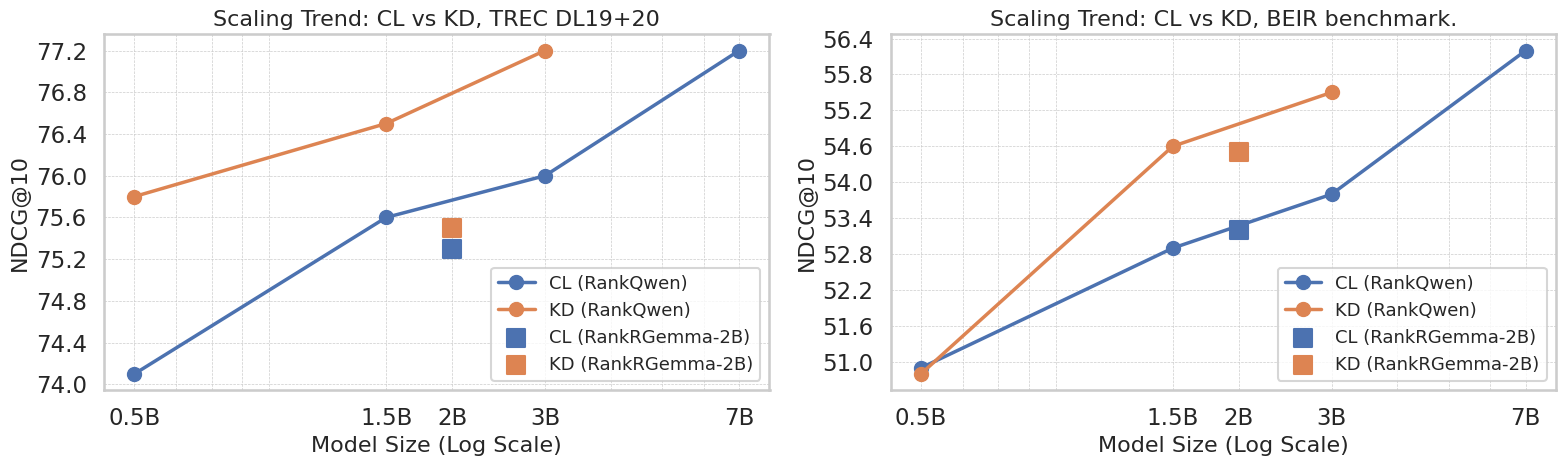}
    \caption{Scaling trend for contrastive learning vs knowledge distillation, for passage reranking. Left figure shows TREC DL19+20 results while right figure shows BEIR results averaged over 12 datasets (excluding Quora). We skip results of RankQwen-7B-KD model.}
    \label{fig:scaling_trend}
    \vspace{-8pt}
\end{figure*}

\paragraph{Scaling model sizes.}
Scaling has been proven effective in IR tasks~\cite[\interalia]{neelakantan2022text,muennighoff2022sgpt,xu2025surveyofmodelarchitectures,zhu2023llmforirsurvey}. We also compare the model size scaling trend between CL versus KD for passage reranking in~\cref{fig:scaling_trend}. We make two observations:
\begin{enumerate}[leftmargin=*]
    \item We observe that contrastive learning demonstrates a clear scaling trend, i.e., performance improves with increased model size when trained on the same data, as also reported by prior works~\cite{ma2023fine,neelakantan2022text,muennighoff2022sgpt,zhuang-etal-2024-promptreps,xu2024rankmamba,muennighoff2024gritlm}.
    \item We also note that when distilling from a larger, more capable teacher model, knowledge distillation also demonstrates a scaling trend. For example, RankQwen's performance on 12 BEIR datasets (excluding Quora) improves from 50.8 to 54.6, 55.5 when scaling from 0.5B to 1.5B and 3B model sizes. 
\end{enumerate}
We hypothesize that with our knowledge distillation training strategy, the passage reranking performance can be further improved by scaling up the student model's model sizes, and distilling from stronger teacher models.

\begin{table*}[h!]
\vspace{0pt}
\centering
\caption{Results with on BEIR benchmark (including MS MARCO passage Dev) with RLHN training mixture. We report NDCG@10 as the performance metric. We mark best result in each row bold; $\dag$ indicates the overall best result excluding the 7B-sized teacher model.
}
\label{tab:results_rlhn}
\resizebox{0.85\textwidth}{!}{
\begin{tabular}{lrrrrrr}
\toprule
\multicolumn{1}{l|}{\,}
& \multicolumn{1}{l|}{Retriever} 
& \multicolumn{1}{l|}{Teacher} 
& \multicolumn{2}{c|}{Contrastive Learning} 
& \multicolumn{2}{c}{Knowledge Distillation} \\
\multicolumn{1}{l|}{Dataset} 
& \multicolumn{1}{r|}{110M} & \multicolumn{1}{r|}{7B} & \multicolumn{1}{r}{0.5B} & \multicolumn{1}{r|}{3B} & \multicolumn{1}{r}{0.5B} & \multicolumn{1}{r}{3B} \\ 
\midrule
MSMARCO & 42.7 & \textbf{46.0} & \,\,\,\,\,\,\,\,\,\,\,\,\,\,\,\,\,45.5 & 45.6$\dag$ & \,\,\,\,\,\,\,\,\,\,\,\,\,\,\,\,\,45.3 & 45.6$\dag$ \\
Arguana & 44.5 & \textbf{79.1} & 65.2 & 76.7$\dag$ & 63.7 & 75.6 \\
Climate-FEVER & 26.6 & 40.5 & 39.0 & 40.1 & 40.2 & \textbf{42.6}$\dag$ \\
DBPedia & 42.2 & \textbf{54.3} & 49.9 & 52.7 & 50.1 & 53.7$\dag$ \\
FEVER & 85.0 & 94.0 & 93.6 & 93.9 & 93.8 & \textbf{94.3}$\dag$ \\
FiQA & 39.9 & \textbf{55.9} & 46.7 & 55.1$\dag$ & 47.3 & 54.2 \\
HotpotQA & 69.1 & \textbf{84.8} & 82.9 & 84.7$\dag$ & 82.7 & 84.6 \\
NFCorpus & 35.4 & \textbf{42.4} & 37.0 & 41.6$\dag$ & 36.8 & 41.6$\dag$ \\
NQ & 58.2 & \textbf{74.5} & 67.5 & 72.8 & 68.2 & 73.5$\dag$ \\
Quora & 86.6 & 77.6 & 78.3 & 78.6 & \textbf{81.1}$\dag$ & 77.1 \\
SCIDOCS & 18.7 & \textbf{27.1} & 22.5 & 25.7$\dag$ & 22.2 & 25.7$\dag$ \\
SciFact & 71.9 & \textbf{81.9} & 78.6 & 81.3 & 78.6 & 81.6$\dag$ \\
TREC-COVID & 69.5 & 88.3 & 85.5 & \textbf{89.0}$\dag$ & 86.1 & 88.7 \\
Touche-2020 & 26.4 & 32.9 & 33.1 & 32.8 & 35.1 & \textbf{35.3}$\dag$ \\
\midrule
Average & 51.2 & \textbf{62.8} & 58.9 & 62.2 & 59.4 & 62.4$\dag$ \\
Average w/o Quora & 48.5 & \textbf{61.7} & 57.4 & 60.9 & 57.7 & 61.3$\dag$ \\
\bottomrule
\end{tabular}
}

\vspace{0pt}
\end{table*}
\begin{figure*}[htp]
    \centering
    \includegraphics[width=\textwidth]{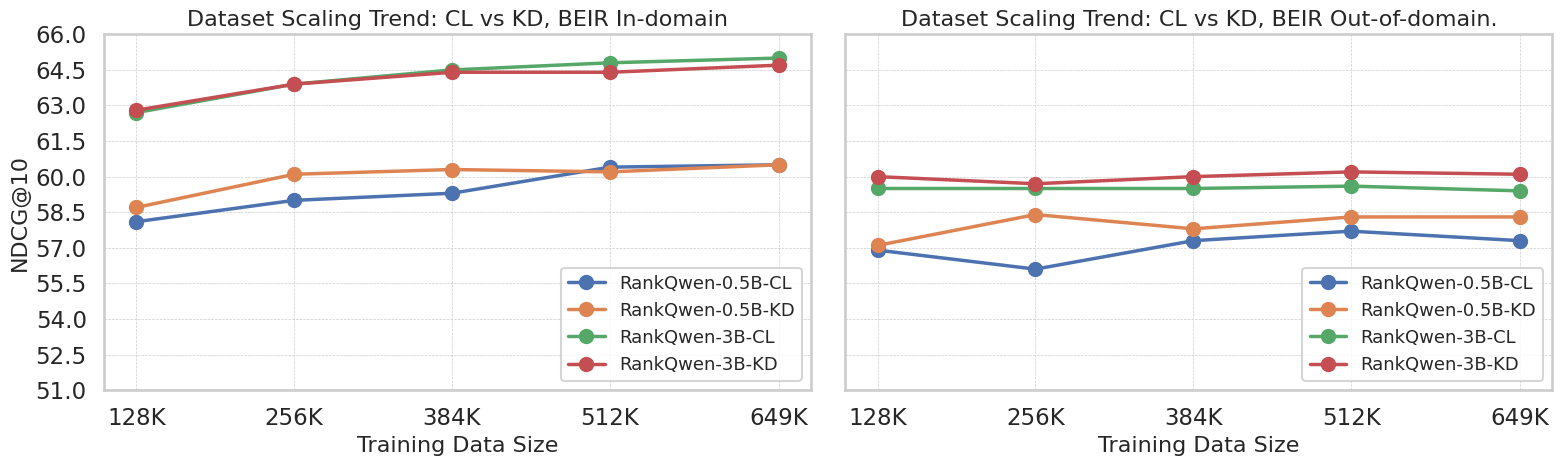}\vspace{-8pt}
    \caption{Dataset scaling trend for CL versus KD. Left figure shows average results for 7 in-domain datasets from BEIR benchmark, while the right figure shows average results for 7 out-of-domain datasets.}
    \label{fig:dataset_scaling_trend}
    \vspace{-8pt}
\end{figure*}
\subsection{Scaling Training Data}
\label{subsec:rlhn_results}
\paragraph{Experiment setup.}
In~\cref{subsec:main_results}, we focus on a controlled experiment with various model sizes. 
Now we examine the efficacy of knowledge distillation when scaling the axis of training data. 
Notice that doing so leads to the decoupling of the retriever and reranker as we can no longer mine hard negatives from the retriever on MS MARCO passage trainset.
We use a recently released trainset, RLHN~\cite{thakur2025fixingdatathathurtsperformance}, which consists of 649K $(q, d^+, D^-)$ training samples pruned from the larger BGE trainset~\cite{bge_embedding}. Specifically, RLHN comprises train samples from the following dataset: MS MARCO passage~\cite{bajaj2016ms}, Arguana~\cite{wachsmuth2018retrieval}, FEVER~\cite{thorne2018fever}, FiQA~\cite{maia201818}, SCIDOCS~\cite{cohan2020specter}, HotpotQA~\cite{yang2018hotpotqa} and NQ~\cite{kwiatkowski-etal-2019-natural}. 
In this case, BEIR benchmark as a whole is no longer considered out-of-domain evaluation.
We use a lightweight \texttt{intfloat/e5-base-v2} retriever~\cite{wang2022e5embedding} pretrained on unlabeled text pairs, then finetuned with the combined MS MARCO passage and NQ training mixture.

We use the similar training strategy as~\cref{sec:experiments} (the only difference is the ranker training data): we first train 7B-scale reranker with Qwen2.5-7B and contrastive learning, then train smaller rerankers with knowledge distillation, with the reranking scores on RLHN dataset labeled by the teacher, as well as the contrastive learning counterparts.
We train 0.5B and 3B rerankers, as we notice Qwen2.5-1.5B fail to converge in the contrastive learning setting. We name the trained rerankers as RankQwen-\{0.5B, 3B, 7B\}-CL-RLHN and RankQwen-\{0.5B, 3B\}-KD-RLHN, respectively.

\paragraph{Results and analysis.}
We report the results in~\cref{tab:results_rlhn}. Compared to~\cref{tab:results_beir}, we notice rerankers trained with RLHN mixture show significant performance boost, suggesting the effectiveness of in-domain training data.

Similar to~\cref{tab:results_beir}, the 7B model achieves the best overall performance, average 62.8 NDCG@10 on 14 BEIR benchmark datasets including MS MARCO Dev. 
We notice that knowledge distillation still outperforms contrastive learning when distilling from the strong 7B teacher model, though the margin is small (<1\%). 
This observation suggests the robustness of knowledge distillation: it can achieve performance improvement compared to contrastive learning when the reranker training data is not coupled with the retriever. 

In ~\cref{fig:dataset_scaling_trend} we further analyze BEIR in-domain (7 datasets used in training) and out-of-domain (7 unseen datasets) performance with varying training data sizes. As the amount of training data increases, both CL and KD show improved in-domain performance, with CL eventually outperforming KD. In contrast, KD consistently outperforms CL on the OOD datasets, although the benefits of increasing training data are diminishing. We leave a more in-depth investigation of the in-domain versus OOD performance gap to future work.

\section{Conclusion and Future Work}
In this paper, we compared the effectiveness of two training strategies, i.e., contrastive learning and knowledge distillation in the context of training text rerankers. With a rigorously controlled experimental setup, we find that when distilling from a larger, more capable teacher model, rerankers trained with knowledge distillation achieve better in-domain and out-of-domain reranking performances compared to contrastive training strategy, and the observation is consistent across different model scales and language model architectures. Along this direction, our future work will investigate a more optimized way to combine two training strategies, and to improve the robustness and out-of-domain generalization of knowledge distillation training. 

Recent works have highlighted the efficacy of a new knowledge distillation paradigm\,---\,on-policy distillation~\cite{agarwal2024policy,yang2025qwen3technicalreport,lu2025onpolicydistillation}. Mapping to the classical information retrieval tasks like reranking, we note that existing training strategies could be generally considered off-policy from the reinforcement learning perspective, where the model is optimized to imitate outputs from external source, either from click information~\cite{joachims2017accurately,tu2025reinforcementlearningrankusing}, human annotation~\cite{bajaj2016ms} or LLMs~\cite{zhang2025qwen3embedding}, which may encounter the train-inference distribution mismatch problem described by~\citet{agarwal2024policy}, and can be traced back to extensive RL literature. We believe a promising future direction is to study how reranking can be improved by on-policy training, where the model gets rewarded based on its current policy.

\section*{Limitations and Potential Risks}
Given the limited computational resources, we are unable to scale the reranker training to >10B models such as Qwen2.5-14B. The observation made in this work may be subject to change for stronger base models. How to improve robustness and out-of-domain generalization of the knowledge distillation training has been extensively investigated by prior works in other NLP domains~\cite{Utama2020Mindthetradeoff,stacey-rei-2024-distilling,Wang2023Serialcontrastiveknowledgedistillation}, we leave investigation for reranking to future works.
This paper focuses on empirical experiments on public benchmarks. We believe this paper do not incur potential risks.

\section*{Acknowledgements}
This material is based upon work supported in part by NSF DMS-2134223.
This research is supported by the National Artificial Intelligence Research Resource (NAIRR) Pilot and the Delta advanced computing and data resource which is supported by the National Science Foundation (award NSF-OAC 2005572).
Any opinions, findings, and conclusions or recommendations expressed in this material are those of the authors and do not necessarily reflect the views of the sponsers.

\bibliography{custom,anthology}

\appendix

\section{Dataset Artifacts and Licenses}
\label{asec:licenses}

Four of the datasets we used in experiments (NFCorpus~\cite{boteva2016full}, FiQA-2018~\cite{maia201818}, Quora\footnote{\url{https://www.kaggle.com/c/quora-question-pairs}}, Climate-Fever~\cite{diggelmann2020climate}) do not report the dataset license in the paper or a repository.
For the rest of the datasets, we list their licenses below:
\begin{itemize}
    \item MS MARCO~\cite{bajaj2016ms}: MIT License for non-commercial research purposes.
    \item ArguAna~\cite{wachsmuth2018retrieval}: CC BY 4.0 license.
    \item DBPedia~\cite{hasibi2017dbpedia}: CC BY-SA 3.0 license.
    \item FEVER~\cite{thorne2018fever}: CC BY-SA 3.0 license.
    \item HotpotQA~\cite{yang2018hotpotqa}: CC BY-SA 4.0 license.
    \item NQ~\cite{kwiatkowski-etal-2019-natural}: CC BY-SA 3.0 license.
    \item SCIDOCS~\cite{cohan2020specter}: GNU General Public License v3.0 license.
    \item SciFact~\cite{wadden2020fact}: CC BY-NC 2.0 license.
    \item TREC-COVID~\cite{voorhees2021trec}: "Dataset License Agreement".
    \item Touche-2020~\cite{bondarenko2020overview}: CC BY 4.0 license.
\end{itemize}


\begin{table*}[h!]
\centering
\resizebox{\linewidth}{!}{%
\begin{tabular}{@{}lrrrrr|rrr@{}}
\toprule
& \multicolumn{1}{c}{BM25} & \multicolumn{1}{c}{GTR-XXL} & \multicolumn{1}{c}{RepLlama2} & \multicolumn{1}{c}{CSPLADE} & \multicolumn{1}{c|}{RepQwen$^\clubsuit$} & \multicolumn{1}{c}{RankT5} & \multicolumn{1}{c}{RankMamba} & \multicolumn{1}{c}{RankLlama2} \\
\multicolumn{1}{l|}{Dataset} & - & 4.8B & 7B & 8B & 7B & 220M & 1.3B & 7B \\ \midrule
\multicolumn{1}{l|}{Arguana} & 39.7 & 54.0 & 48.6 & 48.9 & 54.7 & 33.0 & 34.4 & 56.0 \\
\multicolumn{1}{l|}{Climate-FEVER} & 16.5 & 26.7 & 31.0 & 29.4 & 29.6 & 21.5 & 26.2 & 28.0 \\
\multicolumn{1}{l|}{DBPedia} & 31.8 & 40.8 & 43.7 & 44.5 & 45.2 & 44.2 & 45.8 & 48.3 \\
\multicolumn{1}{l|}{FEVER} & 65.1 & 74.0 & 83.4 & 86.5 & 79.9 & 83.2 & 81.9 & 83.9 \\
\multicolumn{1}{l|}{FiQA} & 23.6 & 46.7 & 45.8 & 40.5 & 45.4 & 44.5 & 43.3 & 46.5 \\
\multicolumn{1}{l|}{HotpotQA} & 63.3 & 59.9 & 68.5 & 69.8 & 68.9 & 71.0 & 76.3 & 75.3 \\
\multicolumn{1}{l|}{NFCorpus} & 32.2 & 34.2 & 37.8 & 37.2 & 38.4 & 38.1 & 39.2 & 30.3 \\
\multicolumn{1}{l|}{NQ} & 30.6 & 56.8 & 62.4 & 60.9 & 62.3 & 61.4 & 52.1 & 66.3 \\
\multicolumn{1}{l|}{Quora} & 78.9 & 89.2 & 86.8 & 87.1 & 87.1 & 83.1 & 83.9 & 85.0 \\
\multicolumn{1}{l|}{SCIDOCS} & 14.9 & 16.1 & 18.1 & 17.6 & 18.3 & 18.1 & 19.6 & 17.8 \\
\multicolumn{1}{l|}{SciFact} & 67.9 & 66.2 & 75.6 & 73.9 & 75.0 & 75.0 & 76.8 & 73.2 \\
\multicolumn{1}{l|}{TREC-COVID} & 59.6 & 50.1 & 84.7 & 83.2 & 85.3 & 80.7 & 79.9 & 85.2 \\
\multicolumn{1}{l|}{Touche-2020} & 44.2 & 25.6 & 30.5 & 38.9 & 36.8 & 44.0 & 37.7 & 40.1 \\ \midrule
Average & 43.7 & 49.3 & 55.1 & 55.3 & 55.9 & 53.7 & 53.6 & 56.6 \\ \bottomrule
\end{tabular}%
}
\caption{Baseline results on BEIR datasets. RepQwen$^\clubsuit$ is a model we trained.}
\label{tab:baseline_results_beir}
\end{table*}
\section{Hyperparameters}
\label{asec:hyperparameters}
For all our training runs, we use a similar set of optimized hyperparameters identified from prior works and our preliminary experiments~\cite{ma2023fine,xu2025csplade}, only ablating learning rate to reduce the effect of overfitting. We use LoRA rank=16 and $\alpha$=32. 
We use AdamW optimizer, learning rate ranging from 3e-5 to 1e-4 with linear warmup and cool down. As we train all models on the train dataset for 1 epoch, we find learning rate is the most important hyperparameter to control overfitting. 
We use 8 GPUs with per device batch 4, gradient accumulation steps 4, which leads to a global batch size of 128. For each $(q_i, d_i^+)$ pair, we use $k=15$ negatives as we find increasing to $31$ negatives may lead to instable contrastive learning. For RepQwen training, we use in-batch negatives; while for reranker training, we focus solely on each $(q_i, d_i^+)$ pair's own hard negatives. 

\section{Baseline Results}
\label{asec:baseline_results}
We report the baseline results in~\cref{tab:baseline_results_beir}.

\end{document}